# Voxel-FPN: multi-scale voxel feature aggregation in 3D object detection from point clouds

Bei Wang[*], Jianping An[*] and Jiayan Cao[*]

Hangzhou Hikvision Digital Technology Co. Ltd

wangbei5, anjianping, caojiayan@hikvision.com

**Abstract:** Object detection in point cloud data is one of the key components in computer vision systems, especially for autonomous driving applications. In this work, we present Voxel-FPN, a novel one-stage 3D object detector that utilizes raw data from LIDAR sensors only. The core framework consists of an encoder network and a corresponding decoder followed by a region proposal network. Encoder extracts multi-scale voxel information in a bottom-up manner while decoder fuses multiple feature maps from various scales in a top-down way. Extensive experiments show that the proposed method has better performance on extracting features from point data and demonstrates its superiority over some baselines on the challenging KITTI-3D benchmark, obtaining good performance on both speed and accuracy in real-world scenarios.

## 1. Introduction

The great success of deep learning network in 2D image detection [1-6] has accelerated the development of 3D object detection techniques. Provided with extra depth information from 3D point cloud though, the difference of data modality between 3D point clouds and 2D RGB images makes it a big challenge in directly transplanting 2D detection techniques. Moreover, with the increase of dimensions and degrees-of-freedom, the objective of predicting exact position, size and orientation in 3D space requires highly-demanding efforts.

In autonomous driving applications, RGB images and 3D point clouds could be simultaneously captured by camera and LIDAR sensors. Using either or both of two modalities, researchers explore effective and reliable solutions for 3D object detection tasks. In terms of representation learning, state-of-the-art work of 3D object detection could be divided into three kinds of methodology in whole: (a) fusion-based approaches, which synchronously fuse region features from RGB images and preprocessed 3D point clouds [7-9]; (b) 2D-detection-driven measures, to conduct subsequent object search in 3D subspace extended from 2D bounding boxes of detection results in RGB images [10]; (c) point-cloud-based methods, exploring the features and inner topology of points to detect 3D objects[11-19].

Currently, fusion-based methods require mature 2D detection frameworks to project the underlying point clouds into bird's eye view (BEV), which might lead to certain degree of information loss. While for 2D-detection–driven approaches, missing objects in RGB image may cause detection failure. Meanwhile, pioneering works [12] on processing raw point clouds have been proposed and further developed to meet the growing demands for higher accuracy on 3D detection.

In this work, we focus on the third issue of aforementioned methodology, to pursue good performance both on speed and accuracy with point cloud only. A novel one-stage 3D detector, named as Voxel-FPN, is proposed. We present main workflow of the 3D detector, which can be viewed a classic encoder-decoder framework that directly operates on point clouds. Unlike existing voxel-based approaches that only utilize points from single scales via a direct forwarding route, we encode multi-scaled voxel features from point data and then aggregated via a bottom-up pathway. Nevertheless, inspired by work [20], a corresponding feature pyramid network (FPN) is designed to decode these feature maps from various scales and associate them with final detection outputs. Experiments indicate that the proposed framework obtains a good balance between speed and accuracy for real-time applications.

---

[*] indicates equal contributions.

In summary, our contribution resides as follows:

1. To the best of our knowledge, it is the first time to conduct multi-scale voxel feature aggregation on point cloud data for 3D detection tasks. We represent a unified and hierarchical way to construct meaningful features representation on point cloud while others [15] prefer to refine 3D proposals extracted in previous stage by utilizing a subnetwork to improve accuracy. Due to the design of bottom-up voxel feature aggregation and top-down pathway of pyramid-like structure, features from multiple scales are jointly considered, thus leading to better performance over other approaches that only extract feature representation at single scale.

2. Extensive search on the appropriate configuration of multi-scale voxel feature aggregation inspires us that when handling point cloud data for 3D detection tasks, "the more, the better" principle does not hold. By conducting experiments on real-world dataset, we identify the most important factors that influence the overall outcomes. A result from KITTI 3D object detection benchmark shows that proposed framework achieves huge increase of mean average precision (*mAP*) on bird's eye view (BEV), ranking the 3rd place and outperforming other approaches. We attribute this mainly to the multi-scale voxel feature aggregation that helps learning more discriminative patterns on BEV plane.

**2. Related Work**

*2.1. Point cloud representation learning methods*

Recent years have witnessed major breakthroughs in point-cloud processing literature. Considering the sparse form of points, 3D object detection approaches make efforts to learn discriminative features from point clouds. A straightforward way is to project 3D points into a 2D plane and utilize traditional 2D convolutional network for feature extraction. To be easy though, projection in reduced dimensionality eliminates space relevance and loses topological information among points.

PointNet [22] applies several non-linear transformations and max-pooling operation on point clouds to predict classification distribution. Due to the shared transforms for each point, the size of learnable parameters is small in practice and the time consumption is relatively low. Moreover, with fully connected layers only affecting on feature dimension, PointNet is irrelevant to the input size of the point sets, allowing point clouds of variable input sizes as input.

When treating each point with shared mapping and exploring global information by observing correlation of point sets entirely, PointNet may omit local features at different local scales. Therefore, PointNet++ [23] focuses on building rich hierarchical features from neighboring points. In work [23], a set abstraction layer consists of sampling local centers, grouping and applying a standard PointNet module to aggregate features of local regions. Set abstraction layers aim to extract features hierarchically and then upsampling operation with interpolation followed by unit PointNet makes output features point-wise.

The advantage of utilizing PointNet is the great flexibility of producing pointwise or global features without requiring any prior knowledge of topology. Along with learned features, points could be further voxelized or grouped as well as the original ones. Such a merit makes PointNet a fundamental component that could be easily embedded in 3D detection frameworks. For example, voxel feature encoding (VFE), proposed in [12], groups points into voxels based on their local coordinates. Similar to PointNet's architecture, linear transforms and max-pooling are conducted in each voxel cells to aggregate features. Voxel-wise features are then collected and transported into middle layers to form descriptive information on the shape of local neighborhood.

*2.2. Fusion based 3D object detection*

MV3D [7] and AVOD [8] project point cloud into multiple views, e.g. in BEV or front view, stack features from corresponding regions in RGB images and finally fuse them to generate a comprehensive representation for detection. Due to misalignment between 2D images and sparsely distributed points in 3D space, bottleneck may exist in such fusion-based methods. Nevertheless, combined efforts should be made to efficiently capture and unify a coupled representation for various data modalities.

*2.3. 2D proposal driven 3D object detection*

F-PointNets extends the 2D detection results into corresponding frustums in the 3D space. Afterwards it employs a PointNet /PointNet++ to segment points into binary (foreground or background) classes and then makes a regression using foreground points. The assumption that only one object exists in a search region is crucial but somehow contradicts with occasions where several objects are closely located and occluded in front view, causing more than one objects in a single frustum. On the other hand, such a detection method relies largely on the quality of the input 2D proposals.

*2.4. Point cloud based 3D object detection*

Methods based on point cloud can be further categorized into two granularity levels, voxel-based and point-based. [12], [13] and [14] belong to the former while PointRCNN represents the latter. [12] groups points into voxels, extracts features in each voxel and aggregates the extracted voxel-wise features in middle convolutional layers for detection. On the observation of sparsity in non-empty voxels, [13] designs sparse convolution algorithm and integrates it to the original framework of [12] to accelerate the calculation of convolutional layers. With the prior knowledge of no overlapping between two objects in height dimension, PointPillars [14] further simplifies SECOND [13] by implementing voxelization only in the BEV plane. As named, points are grouped into vertical columns instead of strided voxels. In comparison with heavy computations of 3D convolution in VoxelNet, PointPillars shifts to 2D convolution, thus greatly reducing the space and time complexity of point feature extraction.

On the other hand, PointRCNN is a two-stage 3D detector. It first extracts point-wise features and regards each point as a regression center for candidate proposals. To reduce overwhelming number of input points, PointRCNN uses standard PointNet++ to segment points in the first stage and only treats foreground ones as regression targets. In the second stage, generated 3D proposals are then gathered with locally aggregated features in region of interest (ROI) for localization and classification refinement.

## 3. Overall architecture

The pipeline of the proposed framework, shown in Fig. 1, consists of three major blocks: voxel feature extraction, multi-scale feature aggregation and a region proposal network with feature pyramid structure (noted as RPN-FPN).

*3.1. Voxel feature extraction*

For feature extraction, 3D spaces are organized into various proposal regions and generate default anchor grids from bird eye view. The 3D space, ranging from $W, H, D$ along the $X, Y, Z$ axes, are divided into equally distributed voxel cells, denoted as $(v_W, v_H, v_D)$, thus producing voxel grids of size $(W/v_W, H/v_H, D/v_D)$. Points are then grouped into those voxels based on individual coordinates $(x, y, z)$. Due to the sparsity of point clouds, point number in each voxels may vary. To reduce learning bias, points are randomly sampled with the same number $N$ for each voxel.

Points in each voxel are collected to form the input feature. Then Voxel Feature Encoding (VFE) [12] blocks, shown in Fig. 2, are utilized to extract voxel-wise features. A VFE block consists of a fully-connected layer, max-pooling and point-wise concatenation to learn pointwise features. After several VFE blocks applied, an element-wise max-pooling operation is conducted to obtain an aggregated output. As a result, input point cloud is transformed into voxel-wise features.

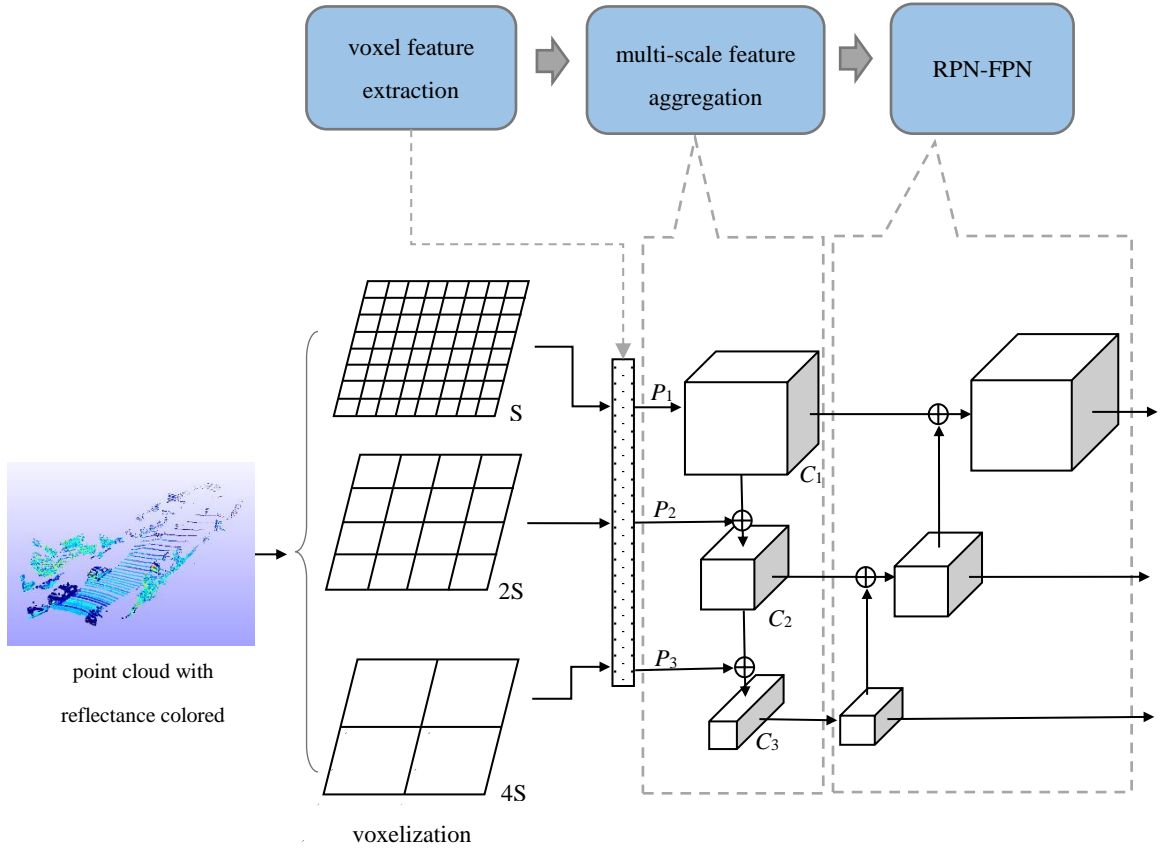

**Figure 1.** Voxel-FPN framework

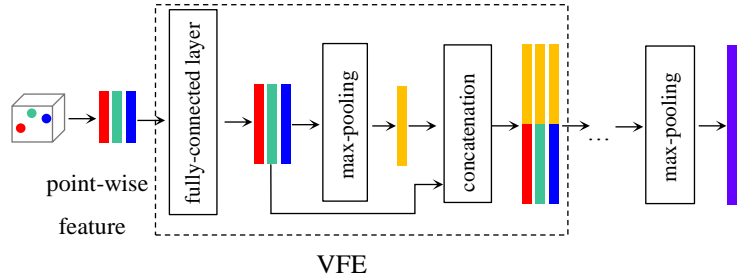

**Figure 2.** Structure of voxel feature extraction network

*3.2. Multi-scale feature aggregation*

Considering point cloud's uneven density across the 3D space, a single setting for default voxel grids can be insufficient to represent all the information available in the scenario. Consequently, we propose a densely aggregated structure to encode voxel feature from multi-scales in a bottom-up way.

Noting a base voxel size as S, we iteratively voxelize a 3D space into multiple sizes of S, 2S, 4S etc., which is given in a serial size of S multiplied by powers of 2. Driven by various combinations of points as input, voxels in multiple scales produce various features. Hence, we expect to aggregate the multi-scale features for better understanding of each voxel cells under bird eye view. Accordingly, we feed feature maps from high resolution into multiple convolutional block with a scaling step of 2. Then, down-sampled outputs are concatenated with voxel feature from coarsen resolution and merged as the input for next scale, thus forming a hierarchy of aggregated features across various voxel scales.

*3.3. Region Proposal Network*

As one of the basic components in modern detectors, region proposal network serves as key module to decode the input feature maps and transform them to candidate boxes. In this work, with the feature maps generated from various voxels of multiple scales as the input to RPN, we adopt Feature Pyramid Network (FPN) [20] design to combine multi-scale features of point cloud. Besides the bottom-up path used in multi-scale feature aggregation, we build a top-down FPN to efficiently construct a rich, hierarchical feature pyramid for multiple voxels features. Each level of the voxel pyramids can be utilized to detect 3D objects in the corresponding regions.

The right of Fig. 1 shows the structure of the RPN-FPN block. With coarser input, we upsample the resolution in x-y plane by a factor of 2. The upsampled maps is then merged with the corresponding map that undergoes 1x1 and 3x3 convolutional layers by element-wise addition. We iterate this process until the finest resolution is met. The final set of voxel features is denoted $\{P_1, P_2, P_3\}$, corresponding to $\{C_1, C_2, C_3\}$ that are respectively of different voxel sizes and ranges. Detection heads are attached on feature pyramid and each default 3D anchors are assigned with corresponding maps, outputting class prediction, location regression and orientation prediction.

*3.4. Loss Design*

Similar to other voxel-based methods in literature, the proposed 3D object detection network utilizes the same localization loss function proposed in [13]. For regression task, the residual targets can be encoded with the following equations:

$$x_t = \frac{x^g - x^a}{d^a}, \ y_t = \frac{y^g - y^a}{d^a}, \ z_t = \frac{z^g - z^a}{h^a}$$
$$w_t = \log \frac{w^g}{w^a}, \ l_t = \log \frac{l^g}{l^a}, \ h_t = \log \frac{h^g}{h^a}, \quad (1)$$
$$\theta_t = \sin(\theta^g - \theta^a)$$

where $x, y, z$ are the center coordinates of the 3D bounding box, and $w, l, h$ are the width, length and height, $\theta$ is the heading angle, and $d^a = \sqrt{(l^a)^2 + (w^a)^2}$; subscripts $t$, $a$ and $g$ indicate the encoded regression targets, the anchor and the labeled bounding box respectively. Thus, total localization loss is given by :

$$L_{\text{loc}} = \sum_{res \in (x,y,z,w,l,h,\theta)} \text{SmoothL1}(\Delta res) \quad (2)$$

For classification branch of the detection output, focal loss is used to handle the imbalance of positive and negative samples:

$$L_{\text{cls}} = -\alpha_a (1 - p_a)^\gamma \log(p_a), \quad (3)$$

where $p_a$ is the probability to be a positive sample for the given class.

The regression scheme mentioned above treats boxes with opposite direction as being the same, so a direction classifier is added to the output of RPN. $L_{\text{dir}}$ term is a softmax classification loss on the discretized directions as in [13]. Altogether, the overall loss is as follows:

$$L_{\text{total}} = \beta_0 L_{\text{cls}} + \frac{1}{N_{pos}}\left(\beta_1 L_{\text{loc}} + \beta_2 L_{\text{dir}}\right), \tag{4}$$

where $\beta_0$=1.0, $\beta_1$=2.0, $\beta_2$=0.2, $N_{pos}$ is the number of positive anchors.

The object detection network is trained for 160 epochs with Adam optimizer, the initial learning rate is 0.0002, and with decay factor 0.8 and a decay period every 15 epochs.

## 4. Experiments

In this section, we focus on extensive experiments and analysis to optimize the network structure and verify the effectiveness of our framework on 3D car detection task.

*4.1. Dataset Description*

All experiment are conducted on KITTI dataset [21], which contains 7481 training and 7518 test pairs of both RGB images and point cloud data. Since the ground-truth annotation are held for benchmark only, we further split the original dataset into 3712 training and 3769 validation samples to verify the effectiveness of the proposed framework [24]. Since objects visible in the image are annotated, we retain point clouds that can be projected to the front camera, which roughly contain 20k points for each scenario.

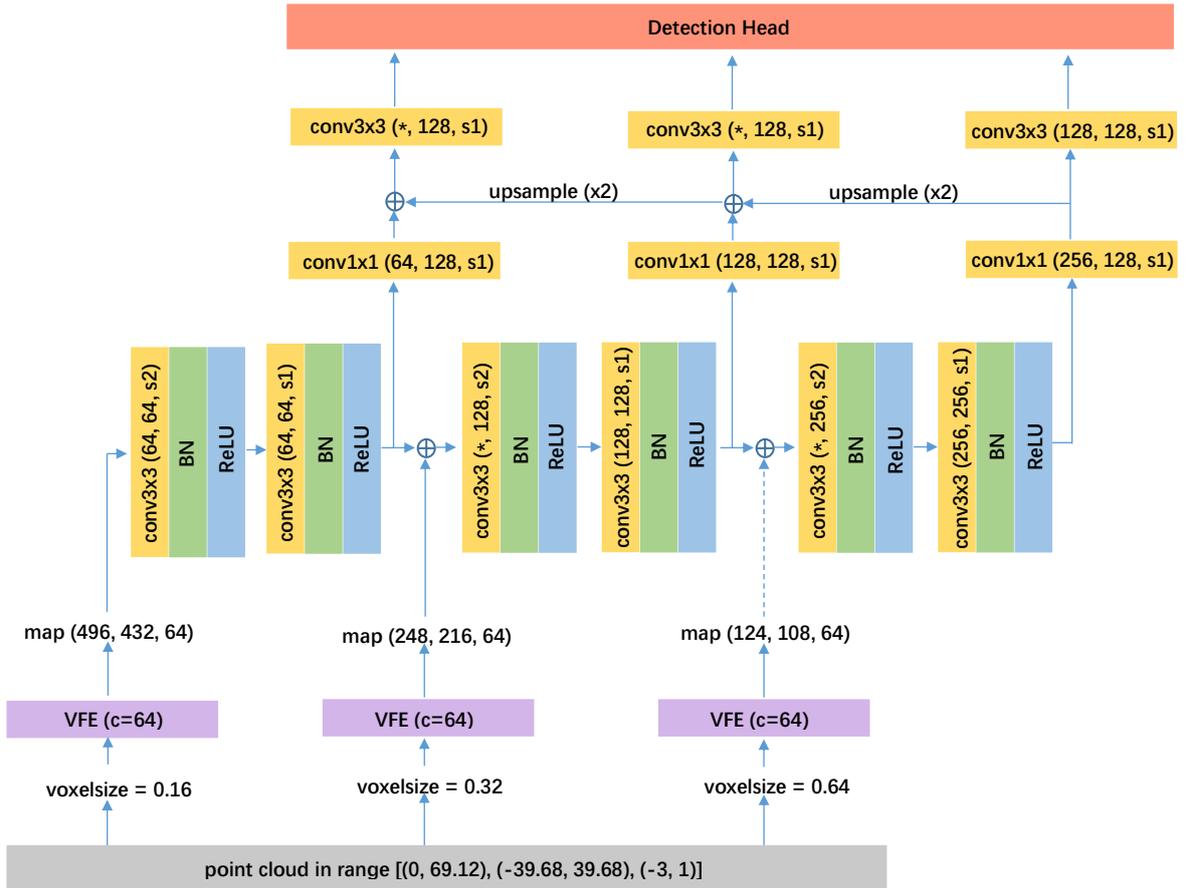

**Figure 3.** The detailed structure for RPN-FPN

*4.2. Implementation details*

In a way similar to PointPillars method, with prior knowledge of no overlapping cars in height, the range of x, y, z in point cloud is restricted to [(0, 69.12), (-39.68, 39.68), (-3, 1)]. Accordingly, three resolution in x-y plane is used to voxelize the point cloud space, producing voxel size of S, 2S, 4S each. Unless otherwise noticed, the implementation settings below is based on S=0.16. The max number of points per voxel is set to [100, 200, 300] and that of voxels is [12000, 8000, 6000]. Then two layers of VFE with 64 output channels are applied in points for voxel feature extraction. After voxelization, point feature $(x_i, y_i, z_i, r_i)$ in each voxel is concatenated with their offset to average point coordinates $(x_i - \bar{x}, y_i - \bar{y}, z_i - \bar{z})$ as VFE input, denoted as $(x_i, y_i, z_i, r_i, l_i, x_i - \bar{x}, y_i - \bar{y}, z_i - \bar{z})$. Therefore, a voxel with point feature of 7 channels is transported into VFE to produce voxel feature of 64 channels. By doing this, we obtain three different voxel feature maps with tensor shapes of (432, 496, 64), (216, 248, 64), and (108, 124, 64), then the largest feature map with a tensor shape (432, 496, 64) becomes the base part of a convolutional network.

We use *conv*(*cout*, *k*, *s*) to represent a standard 2D convolutional block, where *cout*, *k*, *s* denotes output channel number, kernel size and stride respectively. Three convolution blocks are used in RPN, among which block-1 consists of three convolution layers (*conv*(64, 3, 2)-*conv*(64, 3, 1)-*conv*(64, 3, 1)), then five convolution layers (*conv*(128, 3, 2)-*conv*(128, 3, 1)-*conv*(128, 3, 1)-*conv*(128, 3, 1)-*conv*(128, 3, 1)) and then five convolution layers (*conv*(256, 3, 2)-*conv*(256, 3, 1)-*conv*(256, 3, 1)-*conv*(256, 3, 1)-*conv*(256, 3, 1)) are applied to block-2 and block-3 correspondingly. The other two voxel feature maps are concatenated to the corresponding feature maps in the convolutional network with the same shapes.

The detailed implementation of Feature pyramid network in RPN is illustrated in Fig. 3, in which * represent number of output channels depending on the merging strategy among intermediate layers. We compare two forms of merging operations: 'add' or 'concat'. Network utilizing 'concat' operations has better performance than that using 'add' while other configurations are kept the same. In convolutional layers of Fig. 3, $conv\ k \times k(n_1, n_2, s)$ denotes a convolutional module with kernel size $k \times k$, input channels $n_1$, output channels $n_2$ and stride $s$. Since there are only three convolutional layers in the backbone of network, two additional feature maps are produced from VFE before multi-scale voxel feature aggregation. However, in more common cases, design of multi-scale voxel feature aggregation may require more feature maps of different sizes from VFE to be considered, as the same number with convolutional layers in the backbone. Three feature maps are drawn in the Fig. 3 for illustration. In practice, we only use one with voxelsize=0.32 as additional voxel feature, which proved to be better than that with two additional ones in experiments. Finally, detection headers are applied after the last convolution of each block sequentially.

In terms of data augmentation, global flipping, rotation and scaling is applied to the whole point cloud scenarios, where the flipping probability is set to 0.5, the rotation angle ranges from [-π/4, π/4], and the scaling ratio is between [0.95, 1.05]. Followed by [13], during the training process of detection network, several new ground truth boxes and the corresponding points will be put to the same locations of the current scene. Physical collision is tested to ensure only non-overlapping boxes will exist.

*4.3. Ablation Studies*

**Table 1.** Ablation experiments on network components (S=0.16m)

| Multi-scale aggregation | | | RPN-FPN | SSD | 3D detection (*mAP*) | | |
|---|---|---|---|---|---|---|---|
| 1 scale (S) | 2 scales (S+2S) | 3 scales (S+2S+4S) | | | easy | moderate | hard |
| | | √ | × | √ | 86.07 | 76.36 | 69.34 |
| | | √ | √ | × | 85.74 | 75.67 | 69.05 |
| √ | | | √ | √ | 86.56 | 76.47 | 69.43 |
| | √ | | √ | √ | **88.42** | **77.70** | **76.03** |
| | | √ | √ | √ | 86.74 | 77.37 | 75.93 |

Extensive ablation experiments are conducted to probe the contribution of each component. The key components of our framework: multi-scale aggregation and RPN-FPN, are removed respectively to validate 3D detection performance. Table 1 shows the validation results on our pre-defined dataset split for each component's cut-off.

First, to determine the best selection of multi-scale aggregation, we compare three types of combination on voxel inputs. An initial voxel size S=0.16m is found to obtain good performance with an optimized mean average precision (*mAP*). The best *mAP* is multi-scale aggregation with two scales (S+2S). For an initial S=0.16, the best choice of scale amount – two, may stem from that 0.16 and 0.32 voxel sizes enhance feature representation for each other but 0.64 is too large for a mean car size of (1.6, 3.9, 1.56) in width, length and height. It is also concluded that a weak or reductant feature representation, like 0.64 voxel size in our case, may harm the final performance. It shows that for the best performance of a task, not all features should be combined and instead a combing strategy to be searched is of great importance.

Secondly, RPN-FPN is validated by switching FPN structures on and off when given multi-scaled voxel inputs. We achieve better detection outputs on all three category where *mAP* on hard category are greatly increased by 6%.

Last but not the least, we explore the benefits from combing anchor grids on multiple feature maps, thus forming SSD-like output [1]. We simply compare detection outputs from model that adopt the finest feature maps with those that merge with three output feature maps. Marginal gains on overall metrics are obtained due to enlarged sets of anchor grids from various voxel regions.

*4.4. Comparison with other approaches*

Shown in Table 2, we compare performance of our Voxel-FPN method with other approaches in the public KITTI 3D object detection benchmark [25][26]. Our method outperforms for a considerably low runtime. Among these approaches, some top-ranking ones in 3D detection and BEV are selected out and their *mAP*s in moderate level are drawn in Fig. 4 to illustrate the tradeoff between comprehensive performance and time consumption. In *mAP* metric, our method has achieved an excellent performance, especially in moderate level of BEV, which is an important metric for autonomous driving. In comparison with two other voxel-based approaches, i.e. SECOND and PointPillars, our method has better performance. We attribute this to feature representation enhanced by multi-scale voxel feature aggregation.

As reported in work [20], the FPN pathway only introduces relatively small burden due to additional convolutional layers. As seen in Fig. 4, blessed with point cloud data input only, the proposed method has less running time compared to most of the top ranking methods. The only faster method among them, i.e. PointPillars, has lower *mAP* than ours. We believe that the simplicity and efficiency of Voxel-FPN will benefit other point cloud based 3D detectors and serve as a better feature extraction module in future fusion based systems.

**Table 2.** Comparison with other approaches

|  | 3D detection (*mAP*) | | | BEV | | | Modality | Scheme | FPS |
| --- | --- | --- | --- | --- | --- | --- | --- | --- | --- |
|  | easy | mod | hard | easy | mod | hard |  |  |  |
| FPointNet | 81.20 | 70.39 | 62.19 | 88.70 | 84.00 | 75.33 | LIDAR+RGB | 2-stage | 6 |
| MV3D | 62.35 | 71.09 | 55.12 | 86.02 | 76.90 | 68.49 | LIDAR+RGB | 2-stage | 3 |
| AVOD | 81.94 | 71.88 | 66.38 | 86.80 | 85.44 | 77.73 | LIDAR+RGB | 2-stage | 10 |
| MMF | **86.81** | **76.75** | **68.41** | **89.49** | 87.47 | 79.10 | LIDAR+RGB | 2-stage | 12 |
| PointRCNN | 85.94 | 75.76 | 68.32 | 89.47 | 85.68 | 79.10 | LIDAR | 2-stage | 10 |
| SECOND | 84.04 | 75.38 | 67.36 | 88.72 | 86.60 | 78.51 | LIDAR | 1-stage | 25 |
| PointPillars | 79.05 | 74.99 | 68.30 | 88.35 | 86.10 | **79.83** | LIDAR | 1-stage | 62 |
| ours | 85.48 | 76.14 | 68.05 | 89.20 | **87.63** | 79.23 | LIDAR | 1-stage | 50 |

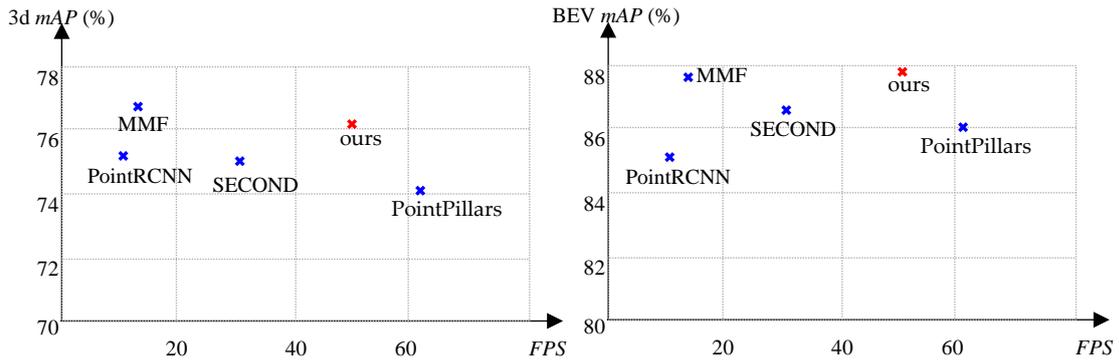

**Figure 4.** *mAP* metric of car 3d and BEV detection in moderate level
compared with some top-ranking approaches

Fig. 5 shows visualized outputs of our proposed framework. We illustrate both projected boxes on RGB image and 3D detection results on point cloud data. In the left column of Fig. 5, a car behind a tree and a white car could be detected out, which is severely occluded and removed from groundtruth annotations in dataset. In the right column of Fig. 5, the car near right boundary is successfully detected, showing the capacity of capturing truncated objects in front eye view.

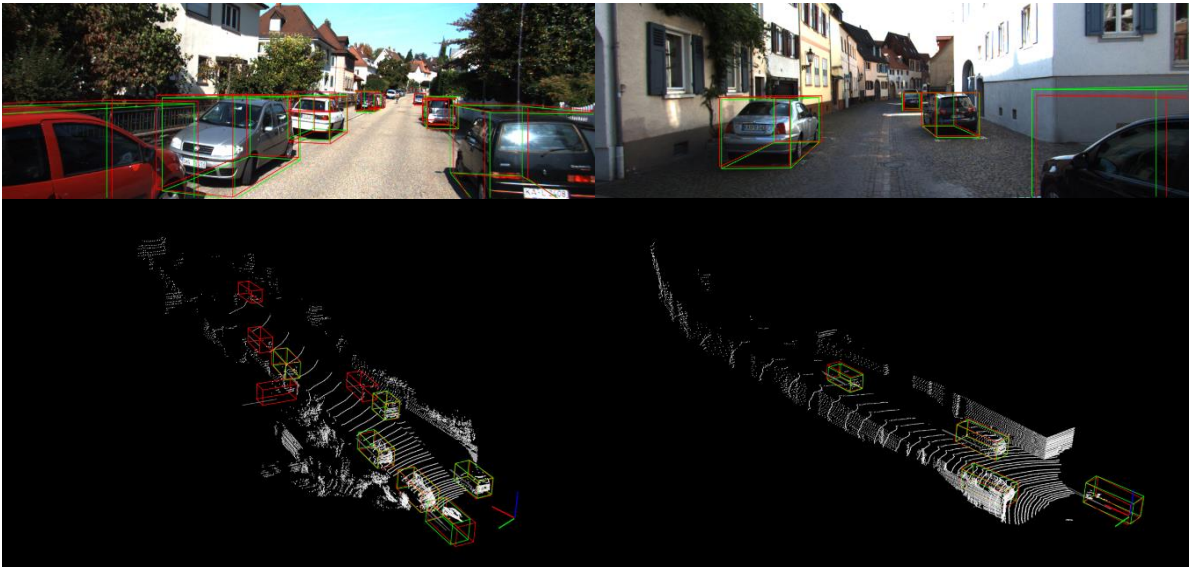

**Figure 5.** Visualized car detection results from our method: cubes in green color denote groundtruth 3D boxes
and those in red indicate detection results

In terms of voxel-based methods, detection results from SECOND, PointPillars and ours are visualized in Fig. 6 for a fair comparison. Under the observation of scenes in validation set, we can inspect that SECOND is more likely to miss objects while PointPillars tends to have more false detections, e.g. in Fig. 6(b). In comparison, results from Voxel-FPN outperforms these two approaches and we attribute this mainly to the enhanced feature presentation by multi-scale voxel feature aggregation.

## 5. Conclusion

In this paper, we propose Voxel-FPN, a novel one-stage 3D detector learned from point cloud data only. It encodes multiple scales of voxel grids from different reception fields and decodes them to final feature maps via a top-down pyramid network, forming a rich hierarchy of feature maps. An appropriate number of aggregating multi-scale voxel features is searched by ablation study. It inspires us that feature representation does not hold the principle "the more, the better" and most important features should be carefully selected to obtain the best performance, remaining a skillful task. Experimental studies indicates that our method is not only competitive in 3D detection outputs but also save great time complexity to be applied in real-world inference tasks.

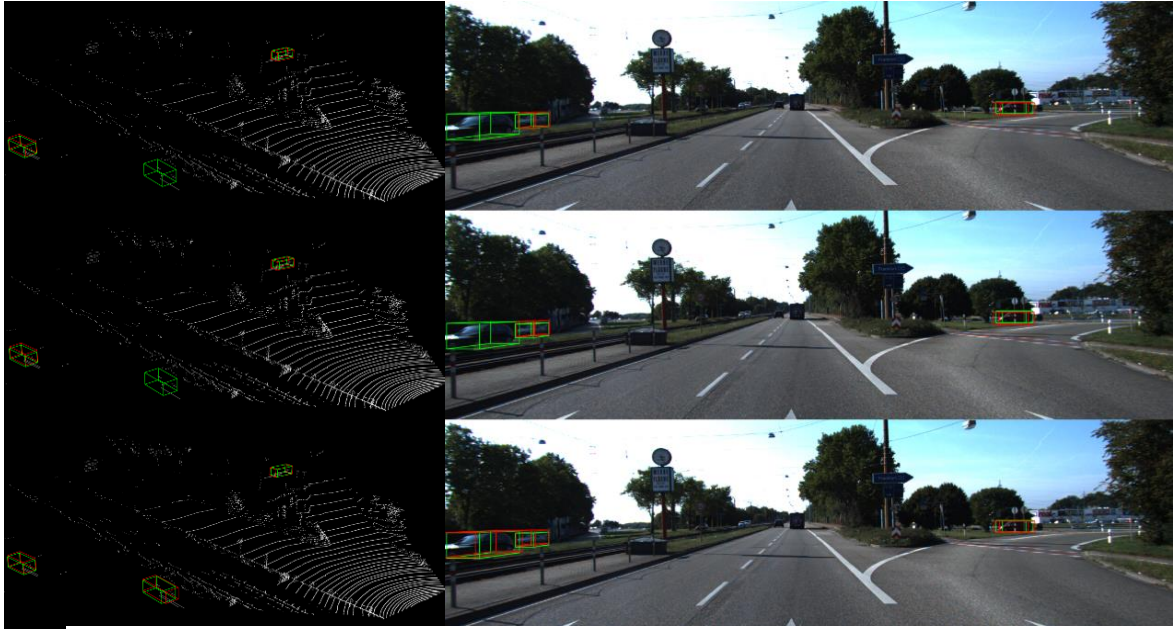

(a)

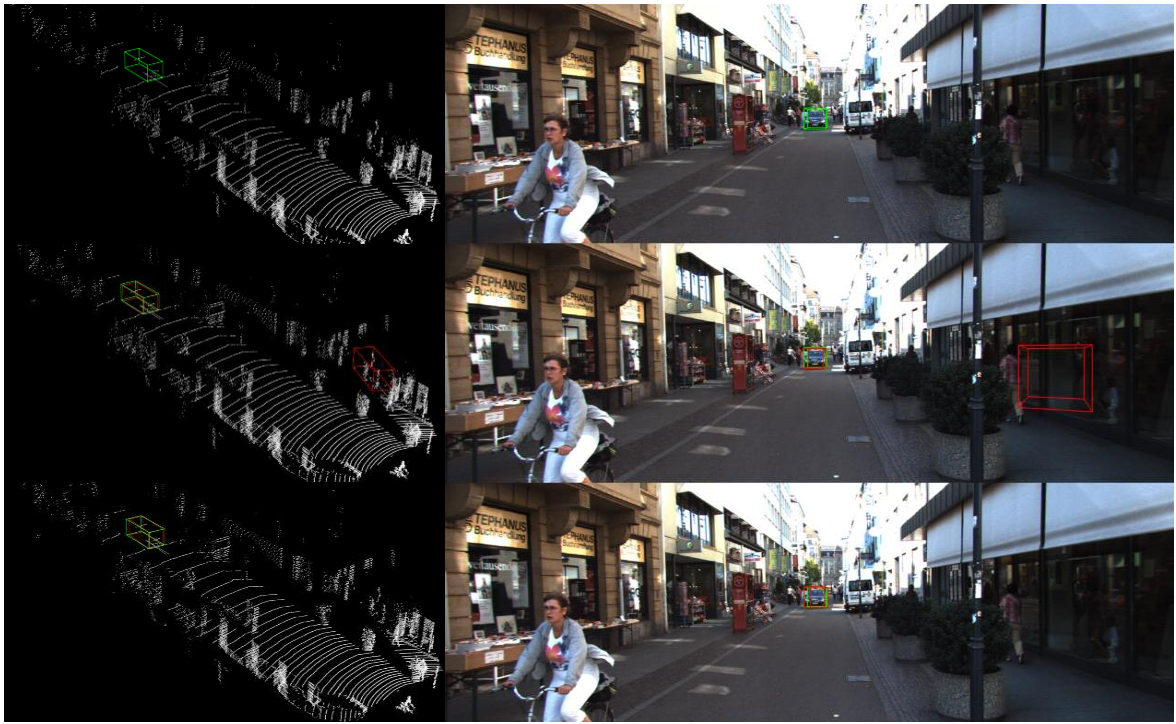

(b)

**Figure 6.** Comparison of results from SECOND (top), PointPillars (middle) and ours (down).